\documentclass{article} 
\usepackage{iclr2025_conference,times}

\iclrfinalcopy 
\usepackage{amsmath,amsfonts,bm}

\def\eqref#1{equation~\ref{#1}}

\def\1{\bm{1}}

\DeclareMathAlphabet{\mathsfit}{\encodingdefault}{\sfdefault}{m}{sl}
\SetMathAlphabet{\mathsfit}{bold}{\encodingdefault}{\sfdefault}{bx}{n}

\usepackage{tikz}
\usepackage{siunitx}
\usepackage{stfloats}
\usepackage{caption}
\usepackage{subcaption}
\usepackage[inline]{enumitem}
\usepackage{amsfonts}       
\usepackage{nicefrac}       
\usepackage{microtype}      
\usepackage{hyperref}
\usepackage{xspace}
\usepackage{mathtools}
\usepackage{titlesec}
\usepackage{booktabs}
\usepackage{relsize}
\usepackage{wrapfig}
\usepackage{float}

\usepackage{amsmath}
\usepackage{amssymb}
\usepackage{mathtools}
\usepackage{amsthm}
\usepackage{bbold}

\theoremstyle{plain}
\theoremstyle{theorem}
\newtheorem{theorem}{Theorem}

\theoremstyle{corollary}

\theoremstyle{definition}
\newtheorem{definition}{Definition}
\theoremstyle{proposition}
\newtheorem{proposition}{Proposition}

\newtheorem*{remark*}{Remark}

\allowdisplaybreaks

\newcommand{\edits}[1]{\textcolor{black}{#1}}
\newcommand{\name}{\textsc{Tracer}\xspace}

\date{}
\author{Alec F. Diallo \\
School of Informatics\\
The University of Edinburgh\\
\texttt{alec.frenn@ed.ac.uk} \\
\And
Vaishak Belle \\
School of Informatics \\
The University of Edinburgh \\
\texttt{vbelle@ed.ac.uk} \\
\And
Paul Patras \\
School of Informatics \\
The University of Edinburgh \\
\texttt{paul.patras@ed.ac.uk}
}

\title{\vspace{-1.5em}Neural Networks Decoded: Targeted and \\Robust Analysis of Neural Network Decisions via Causal Explanations and Reasoning}
\begin{document}
    
\maketitle
       
	\begin{abstract}
    Despite their success and widespread adoption, the opaque nature of deep neural networks (DNNs) continues to hinder trust, especially in critical applications. Current interpretability solutions often yield inconsistent or oversimplified explanations, or require model changes that compromise performance. In this work, we introduce \name, a novel method grounded in causal inference theory designed to estimate the causal dynamics underpinning DNN decisions without altering their architecture or compromising their performance. Our approach systematically intervenes on input features to observe how specific changes propagate through the network, affecting internal activations and final outputs. Based on this analysis, we determine the importance of individual features, and construct a high-level causal map by grouping functionally similar layers into cohesive causal nodes, providing a structured and interpretable view of how different parts of the network influence the decisions. \name further enhances explainability by generating counterfactuals that reveal possible model biases and offer contrastive explanations for misclassifications. Through comprehensive evaluations across diverse datasets, we demonstrate \name's effectiveness over existing methods and show its potential for creating highly compressed yet accurate models, illustrating its dual versatility in both understanding and optimizing DNNs.
    \end{abstract}

	\section{Introduction}\label{sec:introduction}
	Neural networks have demonstrated transformative potential across various applications, notably image classification~\citep{krizhevsky2012imagenet}, medical diagnostics~\citep{esteva2017dermatologist}, and complex pattern recognition~\citep{lecun2015deep}, even surpassing humans in certain domains~\citep{silver2016mastering, rajpurkar2017chexnet}. Yet, their inherent complexity obscures their decision-making processes, turning them into ``black boxes'' that raise transparency and trust concerns, thus impeding their adoption in sectors requiring explainability, such as healthcare and cybersecurity~\citep{zeiler2014visualizing, castelvecchi2016can, doshi2017towards, lipton2018mythos, papernot2018deep, zhang2021understanding}. Neural Network Explainability, pivotal in Explainable AI (XAI), aims to clarify DNN decision-making to ensure trust, ethical application, and bias mitigation. Although various XAI strategies have been proposed, including saliency maps~\citep{zhou2015salient}, Grad-CAM~\citep{selvaraju2017grad}, LIME~\citep{ribeiro2016should}, and SHAP~\citep{lundberg2017unified}, they often present inconsistencies, over-simplification, or architectural constraints, underscoring an ongoing challenge in DNN understanding~\citep{baehrens2010explain, ba2014deep, rudin2019stop}.
		
	In this paper, we introduce \name, a novel approach based on causal inference theory \citep{pearl2009causality}, to infer the mechanisms through which AI systems process inputs to derive decisions. Recognizing that conventional evaluation metrics based solely on validation datasets may not be indicative of a model's performance in real-world settings and drawing inspiration from Pearl's causal hierarchy, our approach reveals how targeted modifications to input features influence the internal states of neural networks, thereby modelling the underlying causal mechanisms. Specifically, \name frames the explainability of neural networks as a causal discovery and counterfactual inference problem, where we observe and analyze all intermediate and final outputs of a model, given any sample, its generated set of interventions, and its counterfactuals. Through the aggregation of multiple such instances, we provide interpretability to state-of-the-art models without requiring any re-training or architectural changes, thus preserving their performance. In conjunction with an efficient approach for counterfactuals generation, this offers contrastive explanations for misclassified samples, expanding our understanding of not just what happened, but why it happened, and what could have happened under different conditions, thus enabling the identification of potential model blind spots and biases, and addressing the overarching issue of trust. \edits{Our main contributions can be summarized as follows:
    	\begin{itemize}
    		\item We propose \name, a framework for estimating the causal mechanisms underpinning DNN decisions, combined with a conditional counterfactual generation method for identifying failure modes, providing actionable insights for improving classifiers.
    		\item We perform comprehensive evaluations of \name on image and tabular datasets, providing explanations for correct and misclassified samples, while highlighting its effectiveness in discovering the causal maps that describe the key transformation steps involved in decisions.
    		\item We demonstrate \name's versatility in both local and global explainability, as well as its ability to outperform prevalent explanation techniques, identify redundancies in neural network architectures, and aid in the creation of optimized, compressed models.
    	\end{itemize}
    }
	
	\section{Related Work}\label{sec:related_work}
	Various techniques have been developed for DNN interpretability, typically categorized by explainability scope, implementation stage, input/problem types, or output format~\citep{adadi2018peeking, angelov2021explainable, vilone2021classification}. Early endeavours like saliency maps by \citet{zhou2015salient}, Grad-CAM~\citep{selvaraju2017grad} and Layer-wise Relevance Propagation (LRP)~\citep{bach2015pixel} visually highlighted key features in input data. Such visual explanation methods often produce inconsistent or coarse explanations, or require structural model changes, sometimes compromising performance or overlooking individual nuances crucial for true comprehension~\citep{rudin2019stop}. Model-agnostic approaches, such as LIME~\citep{ribeiro2016should} and SHAP~\citep{lundberg2017unified} offer explanations by approximating model decision boundaries. However, these methods potentially face challenges such as resource intensiveness or inconsistencies in local explanations. While some works attempt to simplify complex DNNs to improve their interpretability~\citep{che2016interpretable, frosst2017distilling}, they often compromise on performance, as simpler models cannot always capture the nuances of complex DNNs. \edits{In contrast to the aforementioned methods, rather than merely highlighting influential features, \name estimates the causal dynamics that steer DNN decisions, without the need for altering the model or compromising its performance.}
		
			Different from associative methods, causal inference techniques probe deeper, seeking to both understand statistical correlations and uncover the true cause-effect relationships between variables. The idea of merging causal inference with AI is an emerging perspective, advocating for a more robust form of explainability. {rior works on causal inference for AI have primarily revolved around the use of causal diagrams and structural equation models to gain such associative understanding~\citep{pearl2009causality, yang2019causal, xia2021causal, kenny2021post, chou2022counterfactuals, geiger2022inducing, kelly2023you}. 
			\edits{For instance, methods such as those proposed by \citet{kommiya2021towards} and \citet{chockler2024explaining} perform causal reasoning to explain decisions made by image classifiers, focusing on identifying causal elements in the input space, whereas \name extends causal reasoning deeper into the structure of DNNs themselves. By combining causal analysis of both the model's internal workings and the input-output relationships, our approach enables explainabibilty at both the feature and network-structure level, providing more comprehensive explanations for DNN behavior.}
			
			\edits{Recent advances in explainability have also emphasized the importance of counterfactual explanations~\citep{feder2021causalm}, which generate hypothetical instances showing how small changes in inputs would alter a model's prediction. Deep generative models proposed for this task, such as those based on Variational Autoencoders (VAEs)~\citep{pawelczyk2020learning, antoran2020getting} or Generative Adversarial Networks (GANs)~\citep{mirza2014conditional, nemirovsky2022countergan}, typically focus on producing counterfactuals that minimize the required changes to the input features. Our approach improves on these by introducing a dual objective that not only ensures realism through adversarial training but also aligns generated counterfactuals closely with their nearest neighbors in the target class. This guarantees that the counterfactuals generated by \name are both plausible and interpretable.}
			
		Our proposed approach sets itself apart in two main aspects: \edits{(1) rather than only focusing on input features, our approach performs an intervention-based analysis that additionally examines the causal mechanisms within the DNN architecture, identifying how specific layers causally influence the decision-making process, thereby inferring the critical components (critical layers) within DNNs; and (2) a conditional counterfactual generation method, which synthesizes realistic alternative scenarios to identify model blind spots and biases, while ensuring the generated counterfactuals remain plausible and target specific outcomes through controlled feature changes.}

	\section{Theoretical Foundations and Methodology}\label{sec:methodology}
		To understand the internal-workings of DNN architectures, we must consider not only the operations performed by individual layers, but also how they influence one another across the network. \name aims to estimate an accurate model of these mechanisms, focusing on the dynamics that govern the network's decisions. Therefore, our methodology, depicted in Figure~\ref{fig:tracer_overview}, is structured around:
		
		\begin{figure*}[t!]
			\centering
        	\includegraphics[width=.75\linewidth]{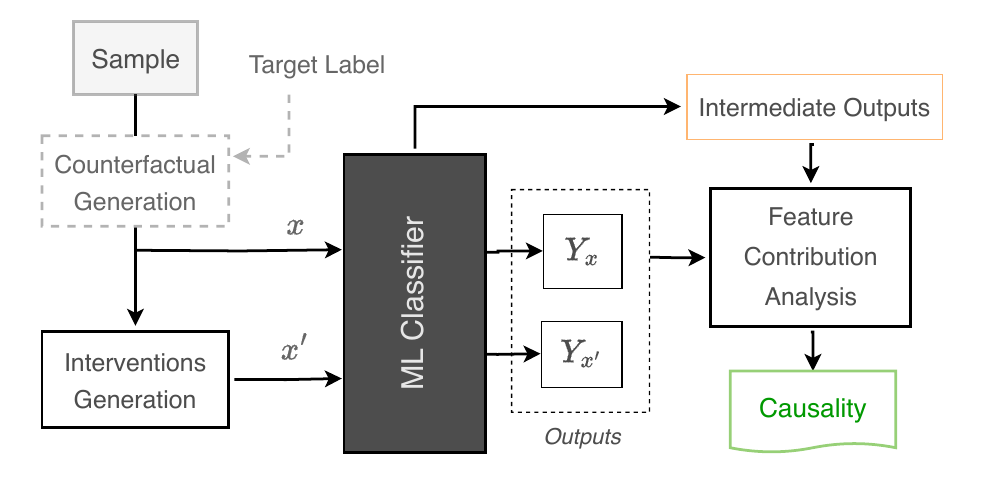}
       		\caption{Overview of \name. Interventions and counterfactuals are used to determine the effects of individual features on the models' intermediate and final outputs, leading to the discovery of the mechanisms underpinning the decision-making process.}
       		\label{fig:tracer_overview}
   		\end{figure*}
   		
		\textbf{Causal discovery:} We analyze the interactions and dependencies within DNNs by systematically altering input features to observe the resulting changes, enabling an effective mapping of the decision pathways. Through this process, we estimate the causal structures that drive the network's decisions, providing a clear understanding of how different features and layers contribute to the outcome.
		
		\textbf{Counterfactual generation:} We simulate alternative scenarios by introducing targeted changes to \textit{input features}, allowing us to explore `what-if' scenarios and observe how specific changes in inputs can lead to different outcomes, providing further insights into the model's sensitivity and robustness.

		\subsection{Causal Theory}\label{subsec:causal_theory}
			Causal theory provides the means to {model} cause-effect relationships, offering a departure from mere observational statistics to tackle questions about interventions and counterfactuals~\citep{pearl2009causality}. To this end, the language of Structural Causal Models has been proposed to formalize these relationships. 
			
			\begin{definition}[Structural Causal Model]
				A Structural Causal Model (SCM) \(\mathcal{M}\) is a 4-tuple \((U, V, \mathcal{F}, P(U))\), where \(U\) is a set of exogenous variables determined by factors external to the model; \(V = \{V_1, V_2, \ldots, V_n\}\) is a set of endogenous variables, each influenced by variables within the model; \(\mathcal{F} = \{f_1, f_2, \ldots, f_n\}\) is a set of functions, each \(f_i\) mapping a subset of \(U \cup V\) to \(V_i\); and \(P(U)\) is a probability distribution over \(U\). For every endogenous variable \(V_i\), its value is determined by \(V_i = f_i(\text{pa}(V_i), U_i)\), where \(\text{pa}(V_i)\) represents the parents or direct causes of \(V_i\), and \(U_i \subseteq U\).
			\end{definition}
			
			Pearl's Causal Hierarchy (PCH), grounded in SCMs, further refines our understanding by categorizing causal knowledge into three distinct levels, which serve as \name's foundations: \\
			\textbf{1. Association}: We extract dependency structures from the DNN activations and outputs $P(Y^{(i)}\ |\ X)$, where $X$ and $Y^{(i)}$ represent the input and the $i$-th layer's output variables, respectively;\\
			\textbf{2. Intervention}: By selectively manipulating feature values, we estimate the intervention distributions $P(Y^{(i)}\ |\ do(X=x_j))$ to understand the effect of particular features on the final decision; \\ 
			\textbf{3. Counterfactual}: We explore alternative (or hypothetical) input scenarios and compute the counterfactual distributions, $P(Y_{X=x'}^{(i)} \mid X=x)$, which quantifies the model's output distribution if a certain input were set to a particular value, given that we actually observed another input.

			\edits{By systematically identifying how specific input features and intermediate layer activations influence the model’s final predictions, \name provides a unique approach, based on the principles of PCH, for capturing an abstract overview of the distinct computational components driving DNN decisions. This structured approach allows us to produce \textbf{explanations} that clarify both the direct influence of features and how the model’s predictions would change under different input conditions, thus providing a more comprehensive understanding of the decision-making process.} 
			
			\begin{definition}[Explanation]
			\edits{Given a $d$-dimensional input $\mathbf{X} = \mathbf{x} \in \mathbb{R}^d$, an explanation for the output $y$ of a model $\mathcal{F}$ is a masked input $\mathbf{x}_E = \mathbf{x} \odot \mathbf{M} \in \{0, 1\}^d$ for which the following conditions hold:
			\begin{enumerate}[label=C\arabic*]
			    \item \textbf{(Correctness):} The model $\mathcal{F}$, when evaluated on the input $\mathbf{x}$, produces the output $y$.\\ $(\mathcal{F}, \mathbf{x}) \models (\mathbf{X} = \mathbf{x}) \text{ and } \mathcal{F}(\mathbf{x}) = y.$
			    \item \textbf{(Sufficiency):}  
			    There exists a \textit{mask} $\mathbf{M} \in \{0, 1\}^d$ such that the resulting explanation $\mathbf{x}_E = \mathbf{M} \odot \mathbf{x}$ produces the same output as the original input: $(\mathcal{F}, \mathbf{x}_E) \models \mathcal{F}(\mathbf{M} \odot \mathbf{x}) = \mathcal{F}(\mathbf{x}) = y.$
			    This condition ensures that the features selected by $\mathbf{M}$ are sufficient to explain $y$.\\Let a mask $\mathbf{M}'$ be defined such that $\mathbf{M}' \not\subseteq \mathbf{M}$, then $ (\mathcal{F}, \mathbf{x}'_E) \models \mathcal{F}(\mathbf{M}' \odot \mathbf{x}) \neq y.$
			    \item \textbf{(Minimality):}
			    The mask $\mathbf{M}$ is minimal, meaning that no strict subset $\mathbf{M'} \subset \mathbf{M}$ satisfies C2. That is, for every mask $\mathbf{M'} \subset \mathbf{M}$, the masked input $\mathbf{x'}_E = \mathbf{M'} \odot \mathbf{x}$ is insufficient to produce the same output: $\forall \; \mathbf{M'} \subset \mathbf{M}, \quad (\mathcal{F}, \mathbf{x'}_E) \not\models \mathcal{F}(\mathbf{M} \odot \mathbf{x}) = y. $
			    C3 guarantees that $\mathbf{M}$ includes the smallest set of features necessary to explain $y$.
			\end{enumerate}
			Note that \textit{(i)} in this definition, $y$ can be set to any specific label to produce explanations for misclassifications or rare events; and \text{(ii)} partial explanations can be simplified to binary decisions (i.e., whether a feature is relevant or not) when computing feature attributions (defined in Section~\ref{sec:causal_effects}).}
			\end{definition}

		\subsection{Causal Discovery}\label{subsec:causal_discovery}
	       To discover a faithful representation of the causal mechanisms underpinning DNN models, we perform an intervention-based analysis where we systematically change the values of input features and study the effects on a given classifier. Such interventions allow us to prove the models and measure the effects of specific changes on the classifier's representations. By observing the internal states and outputs of the classifier, we can deduce how specific components contribute to the final decisions, offering an understanding of the model's causal structure and enabling the identification of key nodes or connections that highly influence the model's predictions. Furthermore, by collecting the observed effects of all interventions, we establish an abstract causal map to visualize the interplay between different network components, and asses their collective influence on the DNN outputs. Ultimately, these detailed visualizations can potentially be instrumental for debugging or refining classifiers, or for designing more interpretable neural network architectures.
			
				\subsubsection{Interventions}\label{sec:interventions}
					In our analysis, interventions are crucial for isolating and understanding the causal significance of specific input features. Given an input vector $x \in \mathbb{R}^d$, where $d$ denotes the dimensionality of the input space, an intervention is simulated by replacing a subset of $x$ with a predetermined baseline value $b$. For a specified subset of indices $I \subseteq \{1, \dots, d\}$ corresponding to the features under intervention, the intervened features are given by: $x'_i = b \cdot \mathbb{1}_{\{i \in I\}} + x_i \cdot (1 - \mathbb{1}_{\{i \in I\}})$, where $\mathbb{1}_{\{i \in I\}}$ indicates 1 when $i$ is in the set $I$ and 0 otherwise. \edits{Assuming $b$ to be causally independent (e.g., binary mask), all input features, before and after interventions, can be considered exogenous variables in the causal map.}
				    
					\begin{proposition}[Causal Isolation of Intervened Samples]
						\edits{Let $F: \mathcal{X} \rightarrow \mathcal{Y}$ denote the mapping function of a DNN. For any $x \in \mathcal{X}$, $I \subseteq \{1, \ldots, d\}$, and $b \in \mathbb{R}$, the intervened sample $x'$ isolates the causal effect of the features in $I$ on $F$ by setting the values of $x_i, \forall i \in I$ to $b$. (Proof in Appendix~\ref{app:proof_proposition1})}
					\end{proposition}
					
					By performing such interventions, we effectively isolate and examine the causal impact of specific features on the output. Through these controlled perturbations, we can determine which features are causally pivotal for the model's decisions, and measure the depth of their influence.
					
				    The chosen baseline values can carry significant importance in our intervention framework. Much like in cooperative game theory where Shapley values~\citep{shapley1953value} use baselines to understand the contribution of each player by averaging their marginal contributions across all possible coalitions, they have been adapted for interpreting machine learning models~\citep{lundberg2017unified}, inspiring our use of baselines to serve as neutral points of reference. Specifically, in our interventions, the baseline value aim to counteract or neutralize the impacts of the specific features being altered. This allows us to isolate the original input's influence on the output without the bias introduced by those features. By contrasting the results from such intervened input with the original, we gain deeper insights into the causal relationships between input features and model outputs.
			
				\subsubsection{Causal Abstraction}\label{sec:causal_abstraction}
					By systematically collecting intermediate and final outputs of the classifier, given an input sample and its interventions, \name enables a focused comparison of representations across network layers and extrapolates an accurate estimation of the causal dynamics driving the network's decisions. For this analysis, we use Centered Kernel Alignments (CKA), a prevalent approach for quantifying the similarity between high-dimensional embeddings. Let $f_i$ and $f_j$ denote the activations of two distinct layers in a neural network for a set of input samples, and let their respective kernel matrices be defined as $K_i = f_i f_i^T$ and $K_j = f_j f_j^T$. Their CKA similarity can then be obtained as:
					\[
						\text{CKA}(K_i, K_j) = \text{HSIC}(K_i, K_j) / \sqrt{\text{HSIC}(K_i, K_i) \times \text{HSIC}(K_j, K_j)},
					\]
					where \(\text{HSIC}(K_i, K_j)\) is the Hilbert-Schmidt Independence Criterion (HSIC) for the kernel matrices, and given by $\text{HSIC}(K_i, K_j) = (n-1)^{-2}\ \text{Tr}(HK_iHK_j).$
					Here, \(H\) is a centering matrix given by \(H = I - \frac{1}{n} \mathbf{1} \mathbf{1}^T\), with \(n\) being the number of samples, \(I\) the identity matrix, and \(\mathbf{1}\) a vector of ones. \(\text{Tr}(\cdot)\) denotes the trace of a matrix.
					The use of CKA for evaluating representation similarity offers several advantages, including: \textit{(i)} \textbf{Normalization:} CKA scores range from 0 (completely dissimilar) to 1 (identical), allowing straightforward comparison across layers; \textit{(ii)} \textbf{Flexibility:} It accommodates various kernel functions, such as linear or Gaussian, enabling flexibility based on specific requirements of the analysis; and \textit{(iii)} \textbf{Robustness:} The use of kernels allows CKA to operate in a richer feature space, providing a more comprehensive similarity measure. Given these properties, CKA stands out as a suitable choice for our similarity analysis of feature representations.
					
					Upon obtaining the similarity measures, we \textit{establish causality by grouping layers based on their CKA values}, where we create a binary matrix \(\mathcal{B}(K_i, K_j)\), which is defined as \(\mathcal{B}(K_i, K_j) = 1 \text{ if } \text{CKA}(K_i, K_j) \geq 1 - \epsilon, \text{ and } 0 \text{ otherwise}\), with \(\epsilon\) representing a predetermined threshold that defines the maximum acceptable dissimilarity for two layers to be considered alike. For our causal analysis, such similarity suggests that these layers contribute to a shared causal node \edits{representing an endogenous variable and describing a distinct structural equation in our causal model}.
				
				\begin{definition}[Layer Groups]\label{def:causal_nodes}
					Let $F(x) = f_0 \circ \ldots \circ f_k(x)$ denote the compositional form of the neural network classifier, with $f_i$ representing the $i$-th layer of the network. And let $\mathcal{B}$ denote the binary CKA matrix. Two distinct layers $f_i$ and $f_j$ are said to belong to the same layer group if and only if $|i - j|=1$ and $\mathcal{B}(K_i, K_j)=1.$ 
				\end{definition}
					
				\begin{theorem}[\edits{Layer Grouping}]\label{ax:layer_group}
					\edits{Let a sequence of layers $\{f_i, f_{i+1}, \ldots, f_j\}$ within a neural network $F(x)$ be classified under the same Layer Group if $\mathcal{B}(K_i, K_{i+1}) = \mathcal{B}(K_{i+1}, K_{i+2}) = \ldots = \mathcal{B}(K_{j-1}, K_j) = 1$, where $\mathcal{B}(K_i, K_j) = 1$ if $\text{CKA}(K_i, K_j) \geq 1 - \epsilon$. The collective causal influence of this sequence on $F(x)$'s output is encapsulated by a single composite layer $g_{ij}$: $F'(x) = f_0 \circ \ldots \circ g_{ij} \circ \ldots \circ f_k(x)$, where $g_{ij} \equiv f_j \circ f_{j-1} \circ \ldots \circ f_i \equiv f_i$. (Proof in Appendix~\ref{app:proof_thm1})}
				\end{theorem}				
				
				{This definition of ``Layer Groups'' aggregates layers into cohesive groups, where each group estimates a distinct node in the decision mechanism of the network. Through this aggregation, we effectively abstract the composition of layers into single causal nodes when their computations are found to be redundant, allowing for a more streamlined and high-level understanding of the network's processes.}

				\begin{theorem}[\edits{Necessary and Sufficient Conditions for Causal Nodes}]
				\edits{Let $F: \mathbb{R}^n \rightarrow \mathbb{R}^m$ be a DNN defined by composition as $F = f_k \circ \ldots \circ f_1$ where each $f_i: \mathbb{R}^{d_{i-1}} \rightarrow \mathbb{R}^{d_i}$ represents the transformation applied by the \(i\)-th layer, $d_0 = n$, and $d_k = m$. Let $g = g(r) = \ldots = g(s) = \{f_i\}_{i=r}^{s}$ with $1 \leq r < s \leq k$ be a subset of consecutive layers. $g$ constitutes a causal node if and only if \\
$\forall i \in \{r, \ldots, s-1\}, \text{CKA}(K_i, K_{i+1}) \geq 1 - \varepsilon,$ where $K_i$ is the kernel matrix of layer $i$ and $\varepsilon \in (0,1)$ is a predefined similarity threshold. (Proof in Appendix~\ref{app:proof_thm2})
}
				\end{theorem}	
				
				\begin{definition}[Causal Links between Layer Groups]\label{def:causal_links}
					Let $g_a$ and $g_b$ denote two distinct layer groups within a neural network. A causal link between $g_a$ and $g_b$ is established if either they are adjacent, meaning there exists at least one pair of layers $f_i \in g_a$ and $f_j \in g_b$ such that $|i - j| = 1$, or if there is at least one pair of layers $f_i \in g_a$ and $f_j \in g_b$ for which $\mathcal{B}(K_i, K_j) = 1$.
    			\end{definition}
    			
    			\edits{By adopting definitions~\ref{def:causal_nodes} and \ref{def:causal_links}, we capture the internal dependencies of DDNs, leading to the discovery of layer-wise abstractions that describe the structural equations governing our causal model. This enriched perspective allows for more powerful explanatory modelling through better understanding of the interplay between layers, and how they collectively shape the network's decisions. Consequently, our approach offers valuable insights into the high-level causal mechanisms that shape the network's behavior, and allows us to provide an abstract, structured, and interpretable view of the causal dynamics that are intrinsic to its operations.}
				
            \subsection{Estimation of Causal Effects}\label{sec:causal_effects}
			To quantify the causal impact of interventions on the neural network's outputs, we define the \textit{Average Causal Effect} (ACE) to quantitatively capture both the direction and magnitude of the effect caused by altering the input features.
			
			\begin{definition}[Average Causal Effect]
    			Let $g_i^\prime(x) = \text{softmax}(g_i(x))$ and $g_i^\prime(x') = \text{softmax}(g_i(x'))$ denote the normalized outputs of a Layer Group $g_i$ for a given input $x$ and its intervention $x'$. The normalization of these outputs is performed to transform the activation scores into valid probability distributions, with which the Average Causal Effect (ACE) can be defined as the expected value of the product of the signed Kullback-Leibler (KL) divergence between their probability distributions:
    			\[
				    \text{ACE}_i = \mathbb{E}_{P(X)}\Big[|\Delta_x^i| \cdot \text{KL}\big(P(g_i^\prime(x) \; | \; do(X=x')) \; \| \;  P(g_i^\prime(x) \; | \; do(X=x))\big)\Big],
				\]
    			where $\Delta_x^i = g_i^\prime(x) - g_i^\prime(x')$ represents the sign of the change induced by the intervention, and $\text{KL}(\cdot)$ represents the KL divergence quantifying the changes between the probability distributions.
			\end{definition}
			This definition provides a robust estimation of the causal effects, offering a comprehensive view of how specific interventions are reflected within the neural network.
			
			\begin{remark*}
				Any intervention that produces outputs sufficiently similar to those produced by the original input has little to no impact on the Average Causal Effect.\\
    			If the intervention on input $x$ to produce $x'$ results in minimal change in the output of a Layer Group $g_i$, such that $g_i^\prime(x) \approx g_i^\prime(x')$, then with all other features of $x$ remaining untouched, the change induced by $x'$ approaches 0, leading to minimal or negligible contribution. Formally, if $g_i^\prime(x) \approx g_i^\prime(x')$, then:
    			\[
       				\text{KL}\big(P(g_i^\prime(x) \; | \; do(X=x')) \; \| \;  P(g_i^\prime(x) \; | \; do(X=x))\big) \approx 0 \implies \text{CE}_i = 0.
    			\]
			\end{remark*}
                
    		This suggests that interventions which do not substantially alter the output of a Layer Group have a negligible causal impact on the model's output, as measured by the ACE. \edits{Our approach henceforth consists of generating interventions, such that those with no effect according to our definition above, are considered not part of the explanation.}
		
		\subsection{Counterfactual Generation}\label{subsec:counterfactual_generation}
			\edits{To improve classification performance and mitigate model biases, we additionally explain misclassified samples through a \name analysis of counterfactuals, highlighting specific feature changes that should be applied to samples in order to obtain the desired outputs.} Counterfactuals are hypothetical data instances that, if observed, would alter the model's decision. Crafting such instances is challenging due to the constraint that all counterfactuals should be valid and plausible. Therefore, using generative models such as Generative Adversarial Networks (GANs)~\citep{goodfellow2020generative}, we aim to achieve this task by including such constraints into our training process. \edits{Particularly, we propose a novel plausibility constraint, whereby the counterfactual generators are trained using both adversarial training to ensure realism, and a proximity-based regularization term to enforce similarity between the generated counterfactuals and real instances from a target class. This ensures that the counterfactuals are realistic and also require minimal changes to the original data. While our proposed constraint can be adapted to various types of generative models (e.g., VAE, GAN, normalizing flows), the model we discuss hereinafter assumes an autoencoder-based GAN architecture.}
			
			Given an input $x \in \mathbb{R}^d$ and a target output $y^*$, our GAN-based counterfactual generation model is defined \edits{such that the generator uses an encoder function \( E_x \) to map the input \( x \) to a condensed latent representation \( z_x = E_x(x) \). The desired model output \( y^* \), typically an integer label, is transformed into a one-hot encoded vector \( o(y^*) \in \mathbb{R}^k \), where \( k \) is the number of classes, using the Kronecker delta function \( o_i(y^*) = \delta_{iy^*} \) for \( i = 1, \ldots, k \). This latent representation \( z_x \), concatenated with the one-hot encoded target label \( o(y^*) \) to form an augmented latent vector \( z = [z_x; o(y^*)] \), is processed by the decoder \( D \) to generate a counterfactual instance \( x^* = D(z) \). 
To verify the authenticity of the generated counterfactual \( x^* \), the discriminator \( \mathcal{D} \) evaluates whether \( x^* \) appears realistic and plausible by distinguishing between original data samples and those produced by the generator.}
			
			The GAN is optimized using a dual objective: \textit{(1)} ensure the authenticity of the generated counterfactual $x^*$ and \textit{(2)} maximize its similarity with its nearest neighbour $x_{\text{nn}}$ among real samples of its training dataset whose label correspond to some target class. This objective can be seen as a combination of a conventional GAN loss and a proximity measure \( d(x^*, x_{\text{nn}}) \), with \( \lambda \) as the balancing coefficient: $ \mathcal{L} = (1-\lambda)\ \mathcal{L}_{\text{GAN}} + \lambda\ d(x^*, x_{\text{nn}}),$ ensuring that generated counterfactuals remain minimally different from real instances in the target class, thereby preserving plausibility while leading to the desired model prediction. This approach offers a flexible and data-efficient process that aligns the generated counterfactuals closely with the actual data distribution, while conditioning on priors for controlled outputs.
			
			\begin{remark*}
			\edits{The regularization distance $d(x^*, x_{\text{nn}})$, essential for maintaining plausibility, can be implemented using metrics such as $\ell_1$, $\ell_2$, or perceptual loss. By introducing minor perturbations $\delta_i$ to the latent representation $z_x$ before decoding, training with this regularization distance enables the generation of multiple distinct plausible counterfactuals $x^* = D([z_x + \delta_i; o(y^*)])$.} 
			\end{remark*}
			
			
			Employing generative models for counterfactual generation, rather than relying on nearest neighbors during inference, offers several advantages. First and foremost, relying on real data points as counterfactuals would require storing large datasets, potentially leading to memory constraints. This could be particularly problematic in applications where storage is expensive or limited. \edits{To address this, we train the counterfactual generator on a small random subset of the training set (e.g., 10\%), which is afterwards discarded, eliminating the need for storage. Moreover, this allows us to generate plausible, novel counterfactuals on-the-fly, avoiding computational costs and latency associated with dataset searches, while enabling broader exploration of the feature space.}
	
	\section{Experiments}\label{sec:evaluation}
		In this section, we evaluate our proposed explainability method, \name, emphasizing both its causal discovery and its counterfactual analysis facets. We perform our initial experiments using the well-known MNIST~\citep{deng2012mnist} and ImageNet~\citep{deng2009imagenet} datasets, which are standards in image classification tasks, and on the CIC-IDS 2017~\citep{sharafaldin2018toward} network traffic dataset to demonstrate \name's applicability to tabular datasets. The MNIST dataset provides a collection of handwritten digits that is ideal for the scrutiny of our methodology, while the diversity and scale of ImageNet offer a broader context for evaluating our approach's effectiveness across a wide range of image recognition challenges. We use pre-trained AlexNet~\citep{krizhevsky2012imagenet} and ResNet-50~\citep{he2016deep} models as our MNIST and ImageNet classifier, respectively, and design a GAN architecture tailored to our counterfactual generation tasks. 
		
		\begin{wrapfigure}{r}{.45\textwidth}
            \vspace{-1.8em}
        	\centerline{
           		\includegraphics[width=\linewidth]{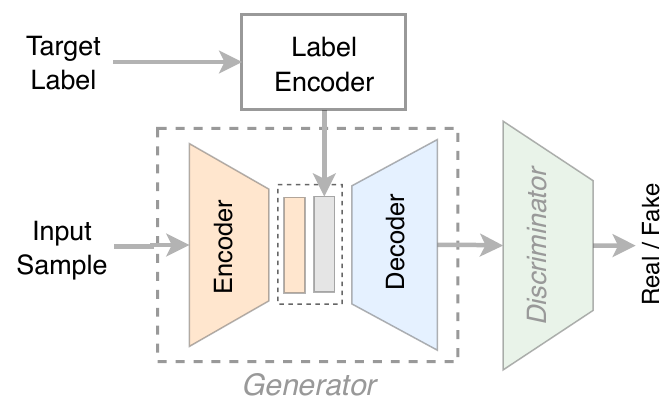}
        	}
       		\vspace*{-0.5em}
       		\caption{Counterfactual GAN architecture.}
       		\label{fig:counterfactual_GAN}
   			\vspace{-1.1em}
   		\end{wrapfigure}   		 
   		 
   		 This GAN, depicted in Figure~\ref{fig:counterfactual_GAN}, consists of a CNN-based Generator for creating plausible, class-conditional counterfactuals, coupled with a CNN-based Discriminator analyzing the authenticity of the generated images, is designed as follows: (1) the encoder uses four convolutional layers to transform the inputs into latent embeddings which are then merged with class information via one-hot label encodings; (2) the decoder uses transposed convolutions to construct the counterfactual inputs from the augmented latent representations produced by the encoder. 
   		 
   		 The Adam optimizer~\citep{kingma2014adam} is used for training, with a learning rate of \(10^{-3}\).
			
   		Through our experiments, we seek to provide a comprehensive understanding of \name's capabilities and the insights it offers into DNN decision-making processes.
		
		\subsection{Causal Discovery and Feature Attributions}
		
		\begin{wrapfigure}{r}{.47\textwidth}
            \vspace{-2em}
        	\centerline{
           		\includegraphics[width=\linewidth]{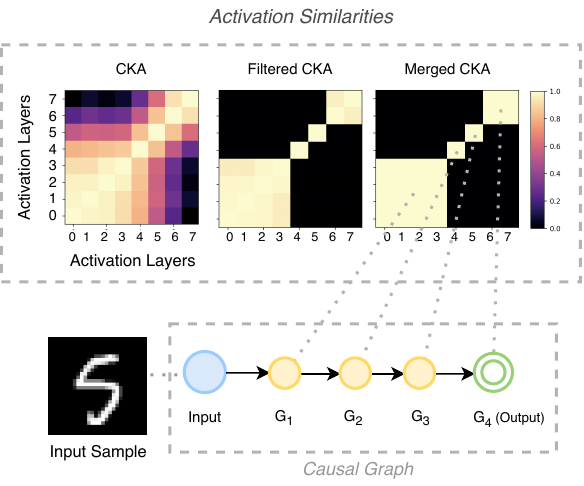}
        	}
       		\caption{\name's causal analysis results for an MNIST sample classified by AlexNet. The causal structure is inferred using CKA similarities between activation outputs from various layers. Nodes in the resulting causal graph symbolize layer groups, while the connections between them capture their causal relationships.}
       		\label{fig:causal_graph_generation}
   			\vspace{-1.3em}
   		\end{wrapfigure}
   		
			To evaluate the effectiveness of \name in uncovering the intricate causal pathways that govern decision-making in DNNs, the relationships between activations of different layers are analyzed using their CKA similarities. Comparing activations produced by the original input and its corresponding interventions illuminates the effect of these interventions on the decisions. As depicted in Figure~\ref{fig:causal_graph_generation}, \name discerns layer groups forming causal nodes and identifies the causal links between them. For instance, eight activation outputs from the AlexNet classifier are observed and analyzed, revealing inherent groupings based on similarity patterns across the network layers. This observation has led to the identification of four distinct causal nodes. Notably, the lack of causal connections between non-adjacent layer groups indicated a \textit{linear causal chain} that informs the network's decision for the analyzed sample.
			
	   		\edits{To quantitatively assess the reliability of \name, we measure how often a given model’s predictions remain consistent when key features identified by our approach are randomly perturbed. \\[.3em] Formally, let $f: \mathcal{X} \rightarrow \mathcal{Y}$ be the classification model, where $\mathcal{X}$ and $\mathcal{Y}$ represent the input and output spaces, respectively. For a dataset $X \subseteq \mathcal{X}$, each sample $x \in X$ is coupled with an explanation mask $M(x)$ generated by an explainability method. Let $P$ denote a perturbation function which modifies $x$ by targeting a proportion $p$ of the significant regions of the explanation. With $x' = x \odot P(M(x), p)$ describing the perturbed sample, the reliability score for the explanations can thus be obtained as: \\[.5em]
			$~~~~ S = |X|^{-1} \sum\nolimits_{x \in X} \mathbb{1}_{\{f(x) \neq f(x')\}},$ where$|X|$ is the number of samples in the dataset, and \\[.5em] $\mathbb{1}_{\{\cdot\}}$ is an indicator function returning 1 if the predictions before and after applying the explainability mask differ, i.e., $f(x) \neq f(x')$.}

			
	   		\begin{wrapfigure}{r}{.45\textwidth}
	            \vspace{-1em}
	        	\centerline{
	           		\includegraphics[width=\linewidth]{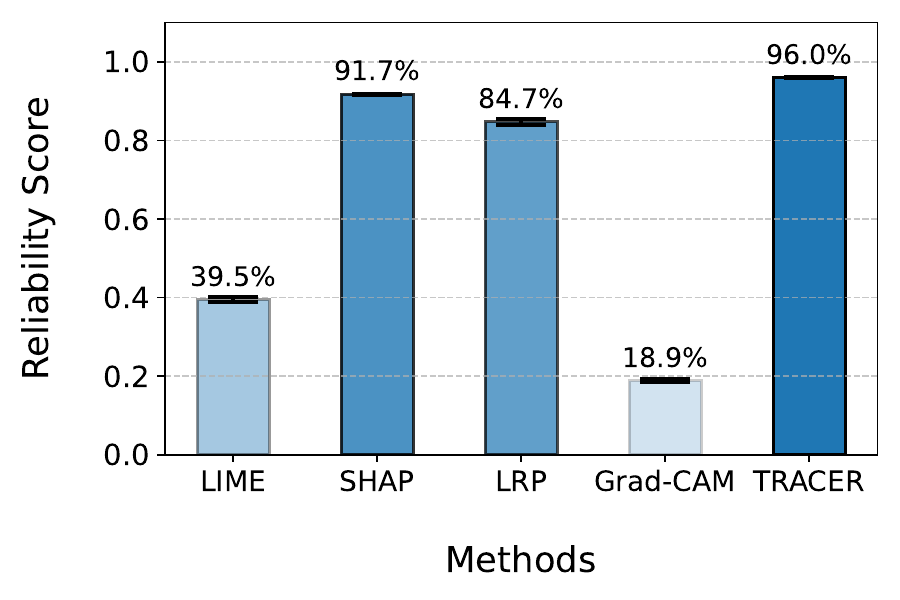}
	        	}
	       		\vspace*{-0.25em}
	       		\caption{\edits{Reliability scores of different explainability methods on the MNIST dataset.}}
	       		\label{fig:reliability_scores}
	   			\vspace{-1em}
	   		\end{wrapfigure}   		
			\edits{This score captures the sensitivity of the model's predictions to changes in areas deemed critical by the explainability method, thereby providing insights into the reliability of the explanations generated. To assess the robustness of \name and compare its performance against that of existing explainability methods, we use this reliability metric on explanations produced by the different approaches when evaluated on all test samples of the MNIST dataset. \\The results, depicted in Figure~\ref{fig:reliability_scores}, show the average and standard deviation of each method's scores over 10 trials, demonstrating \name's superior performance and consistency in producing meaningful and reliable explanations.}
			
		\subsection{Counterfactual Analysis}
			\edits{In this experiment, we explore the use of counterfactuals as a means to understand causes for misclassifications, as well as identify the minimal feature changes required to obtain correct outcomes.}
			\edits{Through a comparison of the causal mechanisms uncovered for the misclassified sample with those for its counterfactuals, \name highlights structural or functional differences resulting from a model's learned parameters, suggesting potential strategies for model improvement, such as refining the training set or implementing regularization techniques.} In essence, counterfactuals offer both an intuitive understanding of model decisions and actionable insights for model enhancement. Appendix~\ref{app:counterfactual_analysis} presents such a counterfactual analysis for a misclassified MNIST sample.
		
			\begin{figure*}[b!]
				\centering
	        	\includegraphics[width=.9\linewidth]{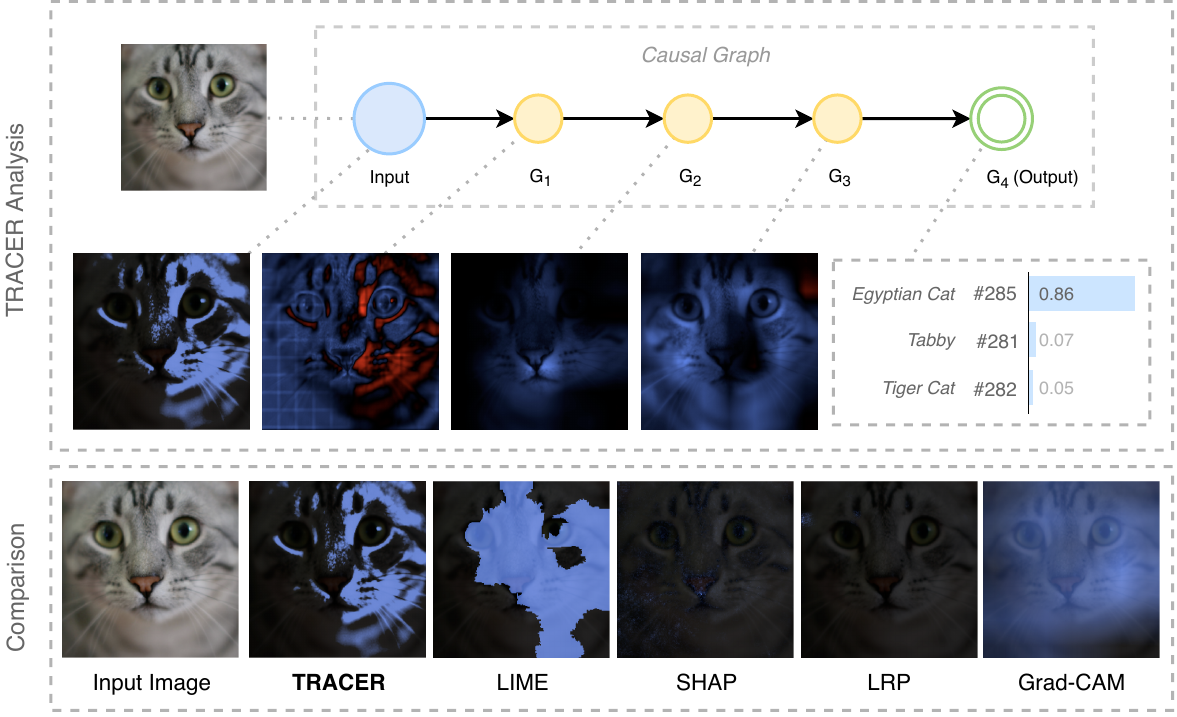}
	       		\caption{\name vs existing XAI methods using an ImageNet sample classified by ResNet-50. The second row shows feature contributions from different causal nodes, while the bottom row compares the explanations provided by different methods. The sparse explanations given by SHAP and LRP may require high-resolution screens for adequate visualization.}
	       		\label{fig:tracer_imagenet}
	       		\vspace{-1em}
	   		\end{figure*}
			
		\subsection{Generalization and Scalability}
			In this experiment, we highlight the broad adaptability of our approach across various neural network architectures and datasets. To this end, we evaluate \name on the ImageNet dataset, as well as on a Network Intrusion Detection dataset, explaining the decisions of both simple and complex NN architectures such as MLP and ResNet-50.
			
		    Given the wide variety and realisic nature of the samples in the ImageNet dataset, its classification results with the ResNet-50 architecture provide a solid benchmark for highlighting the limitations of existing explainability methods and comparing their performances to that of \name. 
		    For this comparison, we selected LIME, SHAP, LRP, and Grad-CAM as benchmarks, since they are among the most widely adopted and representative explainability methods in the literature. The results, depicted in Figure~\ref{fig:tracer_imagenet} show that while existing methods struggle to produce consistent explanations, \name provides coherent and comprehensive explanations that highlight the most important features and patterns that drive the classifier's decisions. Further comparison of these methods, discussed in Appendix~\ref{app:generalization_images}, highlight more distinctions between \name and existing methods, particularly when using DNN architectures that exhibit complex interactions.
		    
			Diving deeper into the versatility spectrum, we challenge \name with the intricacies of structured data using the CIC-IDS 2017 network traffic dataset. This dataset, reflecting authentic network dynamics, unfolds a distinct set of challenges useful for evaluating explainability methods (e.g., diverse data types and intertwined correlations). For example, in an instance where a DDoS-attack-induced traffic is erroneously classified as benign (see Appendix~\ref{app:generalization_ddos}), \name identifies and elucidates features emblematic of the attack through its causal analysis. Specifically, \name reveals that features such as port numbers and data transfer dynamics are essential for the detection of such threats. Overall, the granularity and transparency of explanations provided by \name, especially in domains such as cybersecurity, accentuate its potential to build trust in critical applications.
		
		\subsection{Beyond Local Explainability}
			To evaluate \name's capacity for global explainability, we integrated individual local explanations to form a comprehensive view of a model's decision logic. For this task, we focus on a random subset of the MNIST dataset, processed through the AlexNet architecture, to derive causal insights underpinning the classifier's decisions for all class samples. The results of this analysis, detailed in Appendix~\ref{app:global_explainability}, reveal significant redundancies within AlexNet's architecture for MNIST, allowing us to design compressed representations of the model to optimize the computational efficiency.
			
			The characteristics and comparisons of these compressed models, reported in Table~\ref{tab:model_compression}, show that the most refined model obtained (C1) exhibits a staggering 99.42\% reduction in model size with only a 0.16\% drop in accuracy. This highlights \name's potential for catalyzing practical innovations in DNN design and optimization, without undermining the predictive performance of these models.
			
			\begin{table}[h!]
			    \centering
			    \caption{Comparison of \name-assisted compressed models. $\theta$ represents the number of parameters of the models, and Speed indicates the inference time per sample.}
			    \label{tab:model_compression}
			        \begin{tabular}{r||cccccc}
			            \toprule
			            Model & $\theta$ \textsmaller{\textit{(M)}} & Size \textsmaller{\textit{(MB)}} & FLOPs \textsmaller{\textit{(M)}} & Speed  \textsmaller{\textit{(ms)}} & Accuracy \textsmaller{\textit{(\%)}} \\
			            \midrule\midrule
			            AlexNet & 11.7 & 46.8 & 46.3 & $4.23^{\pm0.4}$ & \textbf{99.64} \\
			            \midrule
			            C3 & 11.5 & 46.3 & 25.0 & $3.21^{\pm0.3}$ & \textbf{99.64} \\[.3em]
			            C2 & 9.4 & 37.9 & 22.9 & $2.65^{\pm0.1}$ & 99.57 \\[.3em]
			            C1 & \textbf{0.06} & \textbf{0.27} & \textbf{13.5} & $\mathbf{1.08^{\pm0.1}}$ & 99.48 \\
			            \bottomrule
			        \end{tabular}
			\end{table}

	\section{Discussions and Limitations}\label{sec:discussion}
			In this study, we focused our evaluations of \name on white-box neural networks. However, its flexibility and design extend beyond, making it equally applicable to black-box models where the internal dynamics remain obscured and only the inputs and outputs are accessible. Under such constraints, \name remains valuable, offering two distinct avenues of exploration. First, it can analyze and quantify the influence of input features on the model's prediction. Alternatively, by using a surrogate white-box model, we can effectively approximate the underlying causal mechanisms driving the predictions. This adaptability underscores \name's potential in diverse environments.
  

			While our \name approach is highly parallelizable by design, its depth of analysis can require a trade-off between granularity (the precision of the causal analysis determined by the number of interventions generated for each sample) and computational efficiency. 
			

	\vspace{-0.5em}
	\section{Conclusion}\label{sec:conclusion}
		In this paper, we introduced \name, a novel approach for accurately estimating the causal dynamics embedded within deep neural networks. Through seamless integration of causal discovery and counterfactual analysis, our methodology enables a deep understanding of the decision-making processes of DNNs. Our empirical results demonstrate \name's ability to both identify the causal nodes and links underpinning a model's decisions, and also leverage counterfactuals to highlight the nuances that drive misclassifications, offering clear and actionable insights for model refinement and robustness. Beyond local explanations, we showcased the potential of our approach to capture the global dynamics of DNNs, leading to practical advantages such as novel and effective model compression strategies. Through our foundational principles and findings, we have ascertained that by producing intuitive, human-interpretable explanations, \name offers outstanding transparency to neural networks, significantly enhancing their trustworthiness for critical applications.

        
	\bibliography{references}

\begin{thebibliography}{48}
\providecommand{\natexlab}[1]{#1}
\providecommand{\url}[1]{\texttt{#1}}
\expandafter\ifx\csname urlstyle\endcsname\relax
  \providecommand{\doi}[1]{doi: #1}\else
  \providecommand{\doi}{doi: \begingroup \urlstyle{rm}\Url}\fi

\bibitem[Adadi \& Berrada(2018)Adadi and Berrada]{adadi2018peeking}
Amina Adadi and Mohammed Berrada.
\newblock Peeking inside the black-box: a survey on explainable artificial intelligence (xai).
\newblock \emph{IEEE access}, 6:\penalty0 52138--52160, 2018.

\bibitem[Angelov et~al.(2021)Angelov, Soares, Jiang, Arnold, and Atkinson]{angelov2021explainable}
Plamen~P Angelov, Eduardo~A Soares, Richard Jiang, Nicholas~I Arnold, and Peter~M Atkinson.
\newblock Explainable artificial intelligence: an analytical review.
\newblock \emph{Wiley Interdisciplinary Reviews: Data Mining and Knowledge Discovery}, 11\penalty0 (5):\penalty0 e1424, 2021.

\bibitem[Antor{\'a}n et~al.(2020)Antor{\'a}n, Bhatt, Adel, Weller, and Hern{\'a}ndez-Lobato]{antoran2020getting}
Javier Antor{\'a}n, Umang Bhatt, Tameem Adel, Adrian Weller, and Jos{\'e}~Miguel Hern{\'a}ndez-Lobato.
\newblock Getting a clue: A method for explaining uncertainty estimates.
\newblock \emph{arXiv preprint arXiv:2006.06848}, 2020.

\bibitem[Ba \& Caruana(2014)Ba and Caruana]{ba2014deep}
Jimmy Ba and Rich Caruana.
\newblock Do deep nets really need to be deep?
\newblock \emph{Advances in neural information processing systems}, 27, 2014.

\bibitem[Bach et~al.(2015)Bach, Binder, Montavon, Klauschen, M{\"u}ller, and Samek]{bach2015pixel}
Sebastian Bach, Alexander Binder, Gr{\'e}goire Montavon, Frederick Klauschen, Klaus-Robert M{\"u}ller, and Wojciech Samek.
\newblock On pixel-wise explanations for non-linear classifier decisions by layer-wise relevance propagation.
\newblock \emph{PloS one}, 10\penalty0 (7):\penalty0 e0130140, 2015.

\bibitem[Baehrens et~al.(2010)Baehrens, Schroeter, Harmeling, Kawanabe, Hansen, and M{\"u}ller]{baehrens2010explain}
David Baehrens, Timon Schroeter, Stefan Harmeling, Motoaki Kawanabe, Katja Hansen, and Klaus-Robert M{\"u}ller.
\newblock How to explain individual classification decisions.
\newblock \emph{The Journal of Machine Learning Research}, 11:\penalty0 1803--1831, 2010.

\bibitem[Castelvecchi(2016)]{castelvecchi2016can}
Davide Castelvecchi.
\newblock Can we open the black box of ai?
\newblock \emph{Nature News}, 538\penalty0 (7623):\penalty0 20, 2016.

\bibitem[Che et~al.(2016)Che, Purushotham, Khemani, and Liu]{che2016interpretable}
Zhengping Che, Sanjay Purushotham, Robinder Khemani, and Yan Liu.
\newblock Interpretable deep models for icu outcome prediction.
\newblock In \emph{AMIA annual symposium proceedings}, volume 2016, pp.\  371. American Medical Informatics Association, 2016.

\bibitem[Chockler \& Halpern(2024)Chockler and Halpern]{chockler2024explaining}
Hana Chockler and Joseph~Y Halpern.
\newblock Explaining image classifiers.
\newblock \emph{arXiv preprint arXiv:2401.13752}, 2024.

\bibitem[Chou et~al.(2022)Chou, Moreira, Bruza, Ouyang, and Jorge]{chou2022counterfactuals}
Yu-Liang Chou, Catarina Moreira, Peter Bruza, Chun Ouyang, and Joaquim Jorge.
\newblock Counterfactuals and causability in explainable artificial intelligence: Theory, algorithms, and applications.
\newblock \emph{Information Fusion}, 81:\penalty0 59--83, 2022.

\bibitem[Deng et~al.(2009)Deng, Dong, Socher, Li, Li, and Fei-Fei]{deng2009imagenet}
Jia Deng, Wei Dong, Richard Socher, Li-Jia Li, Kai Li, and Li~Fei-Fei.
\newblock Imagenet: A large-scale hierarchical image database.
\newblock In \emph{2009 IEEE conference on computer vision and pattern recognition}, pp.\  248--255. Ieee, 2009.

\bibitem[Deng(2012)]{deng2012mnist}
Li~Deng.
\newblock The mnist database of handwritten digit images for machine learning research.
\newblock \emph{IEEE Signal Processing Magazine}, 29\penalty0 (6):\penalty0 141--142, 2012.

\bibitem[Diallo \& Patras(2023)Diallo and Patras]{diallo2023deciphering}
Alec~F Diallo and Paul Patras.
\newblock Deciphering clusters with a deterministic measure of clustering tendency.
\newblock \emph{IEEE Transactions on Knowledge and Data Engineering}, 2023.

\bibitem[Doshi-Velez \& Kim(2017)Doshi-Velez and Kim]{doshi2017towards}
Finale Doshi-Velez and Been Kim.
\newblock Towards a rigorous science of interpretable machine learning.
\newblock \emph{arXiv preprint arXiv:1702.08608}, 2017.

\bibitem[Esteva et~al.(2017)Esteva, Kuprel, Novoa, Ko, Swetter, Blau, and Thrun]{esteva2017dermatologist}
Andre Esteva, Brett Kuprel, Roberto~A Novoa, Justin Ko, Susan~M Swetter, Helen~M Blau, and Sebastian Thrun.
\newblock Dermatologist-level classification of skin cancer with deep neural networks.
\newblock \emph{nature}, 542\penalty0 (7639):\penalty0 115--118, 2017.

\bibitem[Feder et~al.(2021)Feder, Oved, Shalit, and Reichart]{feder2021causalm}
Amir Feder, Nadav Oved, Uri Shalit, and Roi Reichart.
\newblock Causalm: Causal model explanation through counterfactual language models.
\newblock \emph{Computational Linguistics}, 47\penalty0 (2):\penalty0 333--386, 2021.

\bibitem[Frosst \& Hinton(2017)Frosst and Hinton]{frosst2017distilling}
Nicholas Frosst and Geoffrey Hinton.
\newblock Distilling a neural network into a soft decision tree.
\newblock \emph{arXiv preprint arXiv:1711.09784}, 2017.

\bibitem[Geiger et~al.(2022)Geiger, Wu, Lu, Rozner, Kreiss, Icard, Goodman, and Potts]{geiger2022inducing}
Atticus Geiger, Zhengxuan Wu, Hanson Lu, Josh Rozner, Elisa Kreiss, Thomas Icard, Noah Goodman, and Christopher Potts.
\newblock Inducing causal structure for interpretable neural networks.
\newblock In \emph{International Conference on Machine Learning}, pp.\  7324--7338. PMLR, 2022.

\bibitem[Goodfellow et~al.(2020)Goodfellow, Pouget-Abadie, Mirza, Xu, Warde-Farley, Ozair, Courville, and Bengio]{goodfellow2020generative}
Ian Goodfellow, Jean Pouget-Abadie, Mehdi Mirza, Bing Xu, David Warde-Farley, Sherjil Ozair, Aaron Courville, and Yoshua Bengio.
\newblock Generative adversarial networks.
\newblock \emph{Communications of the ACM}, 63\penalty0 (11):\penalty0 139--144, 2020.

\bibitem[He et~al.(2016)He, Zhang, Ren, and Sun]{he2016deep}
Kaiming He, Xiangyu Zhang, Shaoqing Ren, and Jian Sun.
\newblock Deep residual learning for image recognition.
\newblock In \emph{Proceedings of the IEEE conference on computer vision and pattern recognition}, pp.\  770--778, 2016.

\bibitem[Kelly et~al.(2023)Kelly, Chockler, Kroening, Blake, Ramaswamy, Navaratnarajah, and Shivakumar]{kelly2023you}
David~A Kelly, Hana Chockler, Daniel Kroening, Nathan Blake, Aditi Ramaswamy, Melane Navaratnarajah, and Aaditya Shivakumar.
\newblock You only explain once.
\newblock \emph{arXiv preprint arXiv:2311.14081}, 2023.

\bibitem[Kenny et~al.(2021)Kenny, Delaney, Greene, and Keane]{kenny2021post}
Eoin~M Kenny, Eoin~D Delaney, Derek Greene, and Mark~T Keane.
\newblock Post-hoc explanation options for xai in deep learning: The insight centre for data analytics perspective.
\newblock In \emph{Pattern Recognition. ICPR International Workshops and Challenges: Virtual Event, January 10--15, 2021, Proceedings, Part III}, pp.\  20--34. Springer, 2021.

\bibitem[Kingma \& Ba(2014)Kingma and Ba]{kingma2014adam}
Diederik~P Kingma and Jimmy Ba.
\newblock Adam: A method for stochastic optimization.
\newblock \emph{arXiv preprint arXiv:1412.6980}, 2014.

\bibitem[Kommiya~Mothilal et~al.(2021)Kommiya~Mothilal, Mahajan, Tan, and Sharma]{kommiya2021towards}
Ramaravind Kommiya~Mothilal, Divyat Mahajan, Chenhao Tan, and Amit Sharma.
\newblock Towards unifying feature attribution and counterfactual explanations: Different means to the same end.
\newblock In \emph{Proceedings of the 2021 AAAI/ACM Conference on AI, Ethics, and Society}, pp.\  652--663, 2021.

\bibitem[Krizhevsky et~al.(2012)Krizhevsky, Sutskever, and Hinton]{krizhevsky2012imagenet}
Alex Krizhevsky, Ilya Sutskever, and Geoffrey~E Hinton.
\newblock Imagenet classification with deep convolutional neural networks.
\newblock \emph{Advances in neural information processing systems}, 25, 2012.

\bibitem[LeCun et~al.(2015)LeCun, Bengio, and Hinton]{lecun2015deep}
Yann LeCun, Yoshua Bengio, and Geoffrey Hinton.
\newblock Deep learning.
\newblock \emph{nature}, 521\penalty0 (7553):\penalty0 436--444, 2015.

\bibitem[Lipton(2018)]{lipton2018mythos}
Zachary~C Lipton.
\newblock The mythos of model interpretability: In machine learning, the concept of interpretability is both important and slippery.
\newblock \emph{Queue}, 16\penalty0 (3):\penalty0 31--57, 2018.

\bibitem[Lundberg \& Lee(2017)Lundberg and Lee]{lundberg2017unified}
Scott~M Lundberg and Su-In Lee.
\newblock A unified approach to interpreting model predictions.
\newblock \emph{Advances in neural information processing systems}, 30, 2017.

\bibitem[Mirza(2014)]{mirza2014conditional}
Mehdi Mirza.
\newblock Conditional generative adversarial nets.
\newblock \emph{arXiv preprint arXiv:1411.1784}, 2014.

\bibitem[Nemirovsky et~al.(2022)Nemirovsky, Thiebaut, Xu, and Gupta]{nemirovsky2022countergan}
Daniel Nemirovsky, Nicolas Thiebaut, Ye~Xu, and Abhishek Gupta.
\newblock Countergan: Generating counterfactuals for real-time recourse and interpretability using residual gans.
\newblock In \emph{Uncertainty in Artificial Intelligence}, pp.\  1488--1497. PMLR, 2022.

\bibitem[Olvera-L{\'o}pez et~al.(2010)Olvera-L{\'o}pez, Carrasco-Ochoa, Mart{\'\i}nez-Trinidad, and Kittler]{olvera2010review}
J~Arturo Olvera-L{\'o}pez, J~Ariel Carrasco-Ochoa, J~Francisco Mart{\'\i}nez-Trinidad, and Josef Kittler.
\newblock A review of instance selection methods.
\newblock \emph{Artificial Intelligence Review}, 34:\penalty0 133--143, 2010.

\bibitem[Papernot \& McDaniel(2018)Papernot and McDaniel]{papernot2018deep}
Nicolas Papernot and Patrick McDaniel.
\newblock Deep k-nearest neighbors: Towards confident, interpretable and robust deep learning.
\newblock \emph{arXiv preprint arXiv:1803.04765}, 2018.

\bibitem[Pawelczyk et~al.(2020)Pawelczyk, Broelemann, and Kasneci]{pawelczyk2020learning}
Martin Pawelczyk, Klaus Broelemann, and Gjergji Kasneci.
\newblock Learning model-agnostic counterfactual explanations for tabular data.
\newblock In \emph{Proceedings of the web conference 2020}, pp.\  3126--3132, 2020.

\bibitem[Pearl(2009)]{pearl2009causality}
Judea Pearl.
\newblock \emph{Causality}.
\newblock Cambridge university press, 2009.

\bibitem[Rajpurkar et~al.(2017)Rajpurkar, Irvin, Zhu, Yang, Mehta, Duan, Ding, Bagul, Langlotz, Shpanskaya, et~al.]{rajpurkar2017chexnet}
Pranav Rajpurkar, Jeremy Irvin, Kaylie Zhu, Brandon Yang, Hershel Mehta, Tony Duan, Daisy Ding, Aarti Bagul, Curtis Langlotz, Katie Shpanskaya, et~al.
\newblock Chexnet: Radiologist-level pneumonia detection on chest x-rays with deep learning.
\newblock \emph{arXiv preprint arXiv:1711.05225}, 2017.

\bibitem[Ribeiro et~al.(2016)Ribeiro, Singh, and Guestrin]{ribeiro2016should}
Marco~Tulio Ribeiro, Sameer Singh, and Carlos Guestrin.
\newblock " why should i trust you?" explaining the predictions of any classifier.
\newblock In \emph{Proceedings of the 22nd ACM SIGKDD international conference on knowledge discovery and data mining}, pp.\  1135--1144, 2016.

\bibitem[Rudin(2019)]{rudin2019stop}
Cynthia Rudin.
\newblock Stop explaining black box machine learning models for high stakes decisions and use interpretable models instead.
\newblock \emph{Nature machine intelligence}, 1\penalty0 (5):\penalty0 206--215, 2019.

\bibitem[Selvaraju et~al.(2017)Selvaraju, Cogswell, Das, Vedantam, Parikh, and Batra]{selvaraju2017grad}
Ramprasaath~R Selvaraju, Michael Cogswell, Abhishek Das, Ramakrishna Vedantam, Devi Parikh, and Dhruv Batra.
\newblock Grad-cam: Visual explanations from deep networks via gradient-based localization.
\newblock In \emph{Proceedings of the IEEE international conference on computer vision}, pp.\  618--626, 2017.

\bibitem[Settles(2009)]{settles2009active}
Burr Settles.
\newblock Active learning literature survey.
\newblock Computer Sciences Technical Report 1648, University of Wisconsin--Madison, 2009.

\bibitem[Shapley et~al.(1953)]{shapley1953value}
Lloyd~S Shapley et~al.
\newblock A value for n-person games.
\newblock 1953.

\bibitem[Sharafaldin et~al.(2018)Sharafaldin, Lashkari, and Ghorbani]{sharafaldin2018toward}
Iman Sharafaldin, Arash~Habibi Lashkari, and Ali~A Ghorbani.
\newblock Toward generating a new intrusion detection dataset and intrusion traffic characterization.
\newblock \emph{ICISSp}, 1:\penalty0 108--116, 2018.

\bibitem[Silver et~al.(2016)Silver, Huang, Maddison, Guez, Sifre, Van Den~Driessche, Schrittwieser, Antonoglou, Panneershelvam, Lanctot, et~al.]{silver2016mastering}
David Silver, Aja Huang, Chris~J Maddison, Arthur Guez, Laurent Sifre, George Van Den~Driessche, Julian Schrittwieser, Ioannis Antonoglou, Veda Panneershelvam, Marc Lanctot, et~al.
\newblock Mastering the game of go with deep neural networks and tree search.
\newblock \emph{nature}, 529\penalty0 (7587):\penalty0 484--489, 2016.

\bibitem[Vilone \& Longo(2021)Vilone and Longo]{vilone2021classification}
Giulia Vilone and Luca Longo.
\newblock Classification of explainable artificial intelligence methods through their output formats.
\newblock \emph{Machine Learning and Knowledge Extraction}, 3\penalty0 (3):\penalty0 615--661, 2021.

\bibitem[Xia et~al.(2021)Xia, Lee, Bengio, and Bareinboim]{xia2021causal}
Kevin Xia, Kai-Zhan Lee, Yoshua Bengio, and Elias Bareinboim.
\newblock The causal-neural connection: Expressiveness, learnability, and inference.
\newblock \emph{Advances in Neural Information Processing Systems}, 34:\penalty0 10823--10836, 2021.

\bibitem[Yang et~al.(2019)Yang, Liu, Chen, Ma, and Tsai]{yang2019causal}
Chao-Han~Huck Yang, Yi-Chieh Liu, Pin-Yu Chen, Xiaoli Ma, and Yi-Chang~James Tsai.
\newblock When causal intervention meets adversarial examples and image masking for deep neural networks.
\newblock In \emph{2019 IEEE International Conference on Image Processing (ICIP)}, pp.\  3811--3815. IEEE, 2019.

\bibitem[Zeiler \& Fergus(2014)Zeiler and Fergus]{zeiler2014visualizing}
Matthew~D Zeiler and Rob Fergus.
\newblock Visualizing and understanding convolutional networks.
\newblock In \emph{Computer Vision--ECCV 2014: 13th European Conference, Zurich, Switzerland, September 6-12, 2014, Proceedings, Part I 13}, pp.\  818--833. Springer, 2014.

\bibitem[Zhang et~al.(2021)Zhang, Bengio, Hardt, Recht, and Vinyals]{zhang2021understanding}
Chiyuan Zhang, Samy Bengio, Moritz Hardt, Benjamin Recht, and Oriol Vinyals.
\newblock Understanding deep learning (still) requires rethinking generalization.
\newblock \emph{Communications of the ACM}, 64\penalty0 (3):\penalty0 107--115, 2021.

\bibitem[Zhou et~al.(2015)Zhou, Yang, Yuan, Zhou, and Hu]{zhou2015salient}
Li~Zhou, Zhaohui Yang, Qing Yuan, Zongtan Zhou, and Dewen Hu.
\newblock Salient region detection via integrating diffusion-based compactness and local contrast.
\newblock \emph{IEEE Transactions on Image Processing}, 24\penalty0 (11):\penalty0 3308--3320, 2015.

\end{thebibliography}
	\bibliographystyle{iclr2025_conference}
	
\newpage
\appendix
	
	\section{Proofs}
		\subsection{\edits{Proposition 1 \small{[Causal Isolation of Intervened Samples]}}}\label{app:proof_proposition1}
		\begin{proof}
			\edits{By our definition of interventions, $x'$ is derived from $x$ by replacing the features indexed by $I$ with a constant $b$. Assuming this operation to be independent of the original data generation process of $x$ and controlled externally, for the features in $x'$ indexed by $I$, any variation in $F(x')$ with respect to changes in these features can be causally attributed to the intervention itself. Hence, within the scope of our analysis, the causal effect of the features in $I$ is indeed isolated by the intervention.}
		\end{proof}
	
		\subsection{\edits{Theorem 1 \small{[Layer Grouping]}}}\label{app:proof_thm1}
			\begin{proof}
				\edits{Let $F(x) = f_0 \circ f_1 \circ \ldots \circ f_k(X)$ represent a DNN, where $X$ is the input set and each $f_i$ denotes a layer in the network. Given a sequence of consecutive layers $\{f_i, f_{i+1}, \ldots, f_j\}$, let us show that this sequence can be encapsulated into a single composite layer $g_{ij}$ under the condition that these layers belong to the same Layer Group.}
				
				\edits{According to the definition of Layer Groups, the layers $f_i, f_{i+1}, \ldots, f_j$ are grouped together if they exhibit high similarity as measured by their CKA scores. Specifically, the condition $\mathcal{B}(K_i, K_{i+1}) = \mathcal{B}(K_{i+1}, K_{i+2}) = \ldots = \mathcal{B}(K_{j-1}, K_j) = 1$ holds if:}
				
				\edits{\[
				\text{CKA}(K_m, K_{m+1}) \geq 1 - \epsilon, \quad \forall m \in \{i, i+1, \ldots, j-1\},
				\]}
				
				\edits{where $K_m = f_m f_m^T$ is the kernel matrix for the layer $f_m$, and $\epsilon$ is a small threshold allowing minor dissimilarities. Since $\text{CKA}(K_m, K_{m+1}) \approx 1 - \epsilon, \; \forall m \in \{i, i+1, \ldots, j-1\}$, the activations of these layers are highly similar and functionally redundant.}
				
				\edits{Thus, $\{f_i, f_{i+1}, \ldots, f_j\}$ can be treated as performing similar transformations, and represented by a composite layer $g_{ij}(x) = f_j \circ f_{j-1} \circ \ldots \circ f_i(x) \equiv f_m \; \forall m \in \{i, \ldots, j\}$. Therefore, $F(x)$ can be simplified by replacing the sequence $\{f_i, f_{i+1}, \ldots, f_j\}$ with the composite layer $g_{ij}$. The resulting simplified network is:}
				
				\edits{\[
				F'(x) = f_0 \circ \ldots \circ f_{i-1} \circ g_{ij} \circ f_{j+1} \circ \ldots \circ f_k(x) \equiv f_0 \circ \ldots \circ f_{i-1} \circ f_i \circ f_{j+1} \circ \ldots \circ f_k(x).
				\]}
				
				\edits{Thus, the collective causal influence of the layers in $\{f_i, f_{i+1}, \ldots, f_j\}$ on $F(x)$'s output is encapsulated by the single composite layer $g_{ij}$, which we approximate as $f_i$ due to the high similarity between the layers.}
				\end{proof}

	\subsection{\edits{Theorem 2 \small{[Necessary and Sufficient Conditions for Causal Nodes]}}}\label{app:proof_thm2}
	
	\begin{proof}
	\edits{Let $B$ be a binary matrix whose elements $B(K_i, K_j)$ indicate whether the CKA similarity between the activation outputs of layers $i$ and $j$ meet the threshold criterion as defined in Section~\ref{sec:causal_abstraction}.}
	
	\edits{\textbf{(\(\Rightarrow\)) Necessity:} Assume $g = \{f_i\}_{i=r}^{s}$ is a causal node. By definition~\ref{def:causal_nodes}, each layer $f_i$ in $g$ must contribute to the same overarching function that the node represents, i.e., $g = g_i = g_{i+1} \; \forall \; i \in \{r, \ldots, s-1\}$.}
	
	\edits{Suppose, for contradiction, that there exist at least one pair of consecutive layers $(f_i, f_{i+1})$ within the subset $\{f_i\}_{i=r}^{s}$ for which $B(K_i, K_{i+1}) = 0$. This means that $\exists i \in \{r, \ldots, s-1\} \; \text{s.t.} \; \text{CKA}(K_i, K_{i+1}) < 1 - \varepsilon$. By Tracer's definition of a layer group, this indicates a significant change in information content between layers $f_i$ and $f_{i+1}$. Therefore, $f_i$ and $f_{i+1}$ would not be grouped together ($g_i \neq g_{i+1}$), thus contradicting our initial assumption that $g = g_i = g_{i+1}$.}
	
	\edits{\textbf{(\(\Leftarrow\)) Sufficiency:} Conversely, assume:
\[ \forall i \in \{r, \ldots, s-1\}, B(K_i, K_{i+1}) = 1, \]
indicating that each pair of consecutive layers within $\{f_i\}_{i=r}^{s}$ sufficiently maintains the same information content between the layers without significant loss or alteration, i.e., $g_i= g_{i+1}$. Let $Y_i$ and $Y_{i+1}$, respectively define the SCM equations for the causal nodes of layer groups $i$ and $i+1$:}
	\edits{\begin{align*}
	Y_i = \tilde{f}_i(pa(Y_i), U_i), \; Y_{i+1} &= \tilde{f}_{i+1}(pa(Y_{i+1}), U_{i+1}) \\
	 &= \tilde{f}_i(pa(Y_{i+1}), U_{i+1}),
	\end{align*}}
	\hspace*{-0.3em}\edits{where $pa$ denotes the parent sets, and $U$ represents potential exogenous variables or internal noise within the layers. }
	
	\edits{Since $g_i = g_{i+1}$, then it must be that
	\begin{align*} Y_i \approx Y_{i+1} &\Rightarrow \tilde{f}_i(pa(Y_i), U_i) \approx \tilde{f}_i(pa(Y_{i+1}), U_{i+1}) \\
	&\Leftrightarrow \tilde{f}_i(pa(Y_i), U_i) \approx \tilde{f}_i(pa(Y_i), U_{i+1}).
	\end{align*}}

				\edits{It follows that $U_i \approx U_{i+1}$. Hence, under the assumption of causal sufficiency and the observed similarity, we can conclude that $\tilde{f}_i$ and $\tilde{f}_{i+1}$ represent equivalent causal mechanisms. Therefore, $g = \{f_i\}_{i=r}^{s}$ functions as a unified entity, consistent with the properties of a causal node.}

				\edits{($\Leftrightarrow$) Thus, the condition $B(K_i,K_{i+1}) = 1 \; \forall~i~\in~\{r,~\ldots,~s-1\}$ is both necessary and sufficient for $g$ to be identified as a causal node, validating the causal grouping dictated by the similarity threshold $\varepsilon$, as defined by Theorem~\ref{ax:layer_group}.}
	\end{proof}
	
	\section{\edits{Feature Attributions at Causal Nodes}}
		\edits{Figure~\ref{fig:feature_contributions} below depicts how individual features contribute to the network's final decision. For every causal node {(group of neural network layers)}, we highlight the top contributing features (top convolution filter output or top-3 feature outputs for linear layers). Positive contributions are distinctly marked in blue, signifying features that positively influence the network's decision, while negative contributions are depicted in red, pointing out the features that negatively affect the decision.}
		
			\begin{figure}[h!]
			    \centering
			    \includegraphics[width=.55\linewidth]{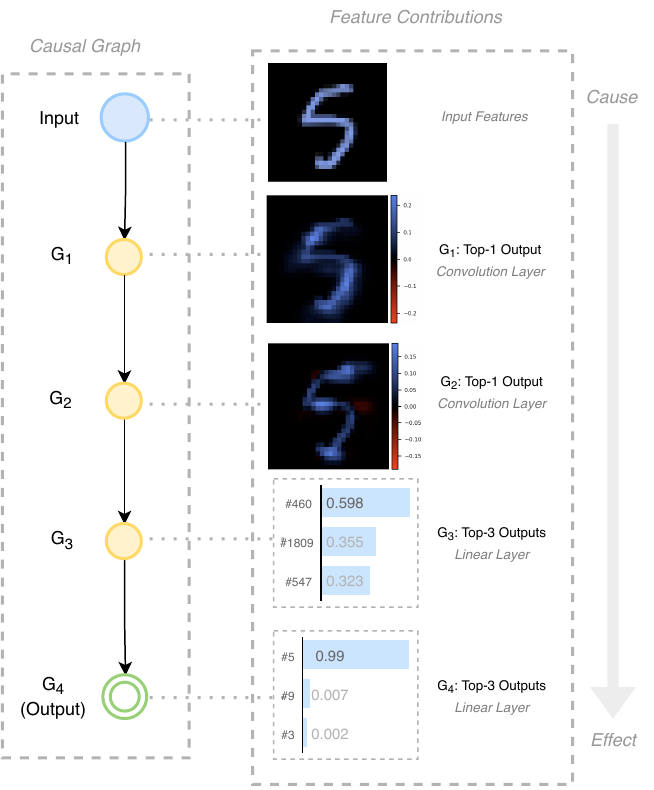}
			    \caption{\edits{Contribution of features within each causal node. Blue and red respectively indicates positive and negative contributions. The overlay on the input sample provides a cohesive visualization of how distinct features of the input affect the final decision via the causal mechanism discovered.
			    }}
			    \label{fig:feature_contributions}
			\end{figure}

	\section{Detailed results: Counterfactual Analysis}\label{app:counterfactual_analysis}
			The objective of counterfactual generation in the context of our research is to offer interpretable insights into the decision-making processes of deep neural networks, particularly in cases of misclassification. By examining the contrast between the original input and the generated counterfactual, we can uncover subtle features or patterns that influence the model's decision, thereby pinpointing what changes might rectify misclassifications.
			
			\begin{figure}[h!]
			    \centering
			    \includegraphics[width=.45\linewidth]{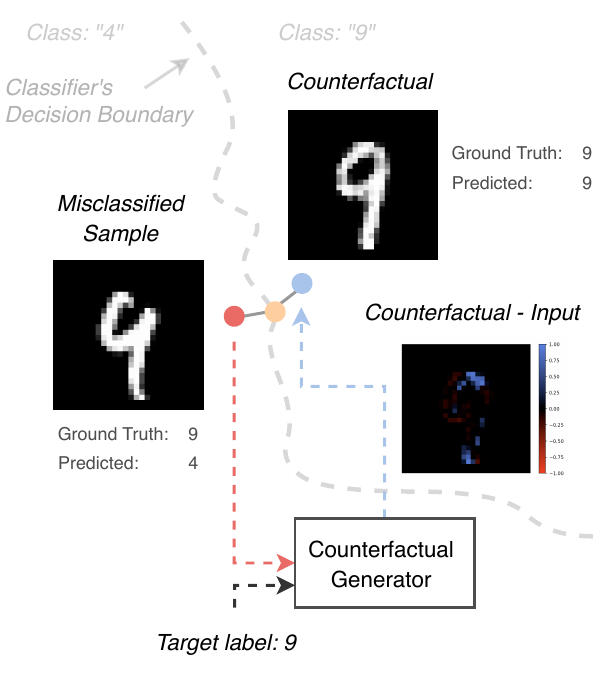}
			    \caption{Illustration of a misclassified MNIST sample and its generated counterfactual. 
			    }
			    \label{fig:counterfactual_generation}
			\end{figure}
			
			As illustrated in Figure~\ref{fig:counterfactual_generation}, given an initially misclassified input and a desired target label, our GAN-based counterfactual generator produces an alternative version of the input, which, when fed to the model, results in the desired outcome. The differences between the input and its counterfactual reveal the minimal modifications required for the classifier to produce the correct (desired) decision.
			
			\begin{figure*}[ht!]
			    \centering
			    \includegraphics[width=\textwidth]{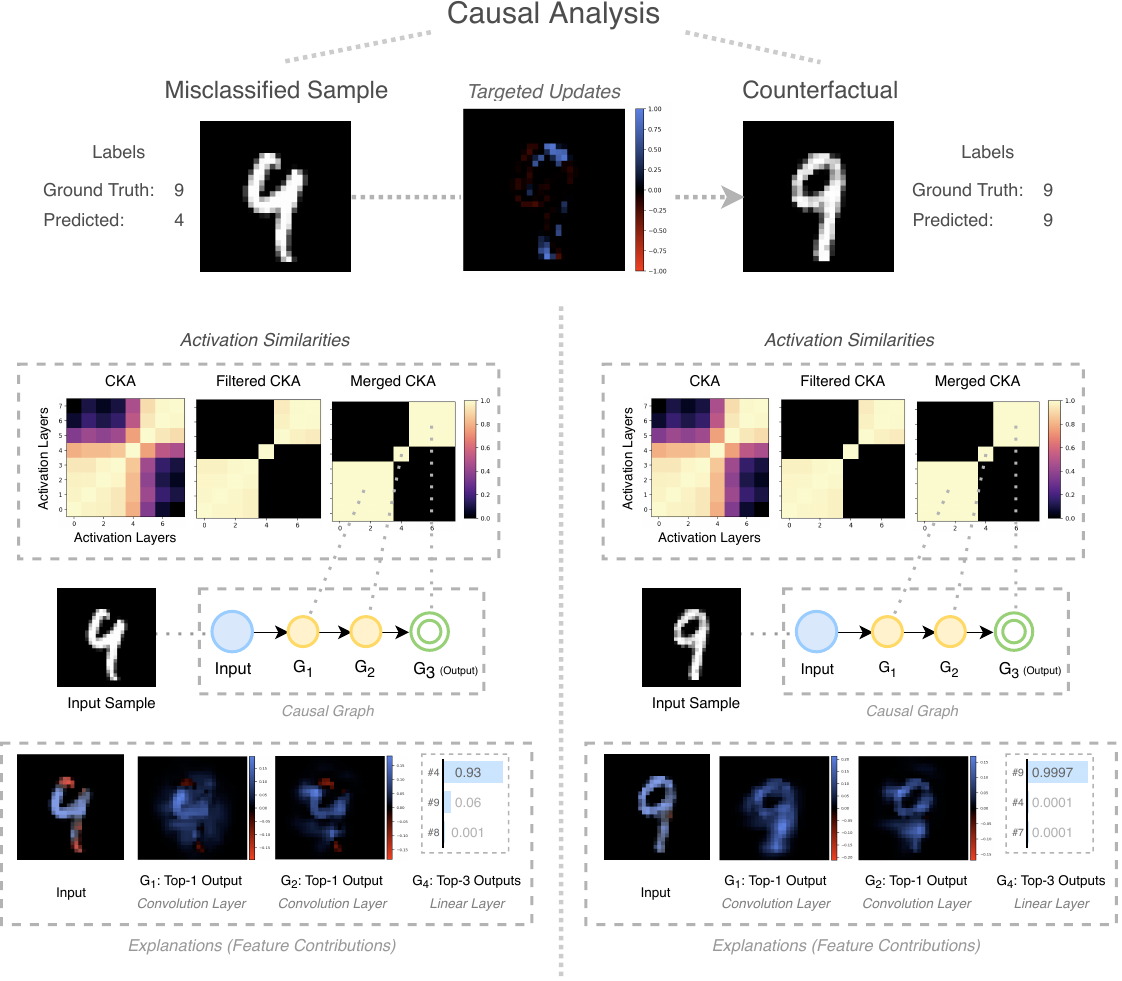}			    \caption{Comparison of an original misclassified input, the generated counterfactual, and the associated causal mechanisms. The variations between the original and counterfactual inputs highlight the pertinent features influencing the model's decision-making process. {(Blue: Positive contributions; Red: Negative contributions; $\text{G}_i$: $i\text{-th}$ Layer Group)}}
			    \label{fig:counterfactual_analysis}
			\end{figure*}

			Through a side-by-side analysis of the causal mechanisms obtained from the predictions of the misclassified sample and its counterfactual, \name provides clear insights into the primary contributing factors to the initial misclassification, while also highlighting via the counterfactual's analysis, the optimal neural pathways for the network to yield the correct (and desired) outcome. This detailed causal analysis is visually represented in Figure~\ref{fig:counterfactual_analysis}. Upon examination, we discern that a predominant portion of the input features, represented in blue, activate neurons that steer the classifier towards the produced outcome in both cases. However, the causal analysis of the misclassified sample reveals a notably more extensive set of features that oppose the predicted outcome when contrasted with the counterfactual. This observation makes it evident that \name not only identifies which parts of the input features support the misclassification (in blue) but also which features contradict this decision (in red). Intriguingly, while the causal graphs remain consistent for both inputs in this instance, the classifier's activations manifest pronounced differences. This insight suggests that the model's learned parameters might lack the flexibility to generalize enough to correctly discern the true label of the misclassified sample. To address this, potential avenues might include incorporating such misclassified instances into the training set or fine-tuning the model with regularization techniques to enhance its generalization capabilities.
			
			This causal analysis reveals that our counterfactual generation method serves two main purposes. First, it provides an intuitive visualization for understanding the nuances of model decisions. Secondly, from a model development and refinement perspective, these counterfactuals can highlight potential vulnerabilities or biases in the model, guiding further training or fine-tuning endeavours.
			
	\section{Detailed results: Generalization}\label{app:generalization}
			
			\subsection{Image Datasets}\label{app:generalization_images}
			Here, we address the question of scalability of \name to large-scale image datasets. Given the challenges associated with the explainability of real-world images (e.g., the intricacies of pixel-level interactions, variances in image quality, or scale), we use for this task the MNIST and ImageNet~\citep{deng2009imagenet} datasets, classified with the AlexNet and ResNet-50 architectures respectively. Using the ImageNet dataset, known for its vastness, diversity, and complexity, we show that \name overcomes the limitations of existing explainability methods.
			The explanations produced by \name and benchmark explainability methods are depicted in Figure~\ref{fig:comparisons}, showing that while existing methods struggle to produce coherent and comprehensive explanations, \name consistently reveals the core features and patterns crucial for classification decisions. The effectiveness of our proposed approach becomes even more apparent when used with complex models like ResNet-50, as it still maintains its precision despite the intricate patterns leveraged by very deep networks, emphasizing its capability to accurately discern the nuances of complex interactions within deeper architectures. 
			
			\begin{figure}[htbp!]
			    \centering
			    \includegraphics[width=\linewidth]{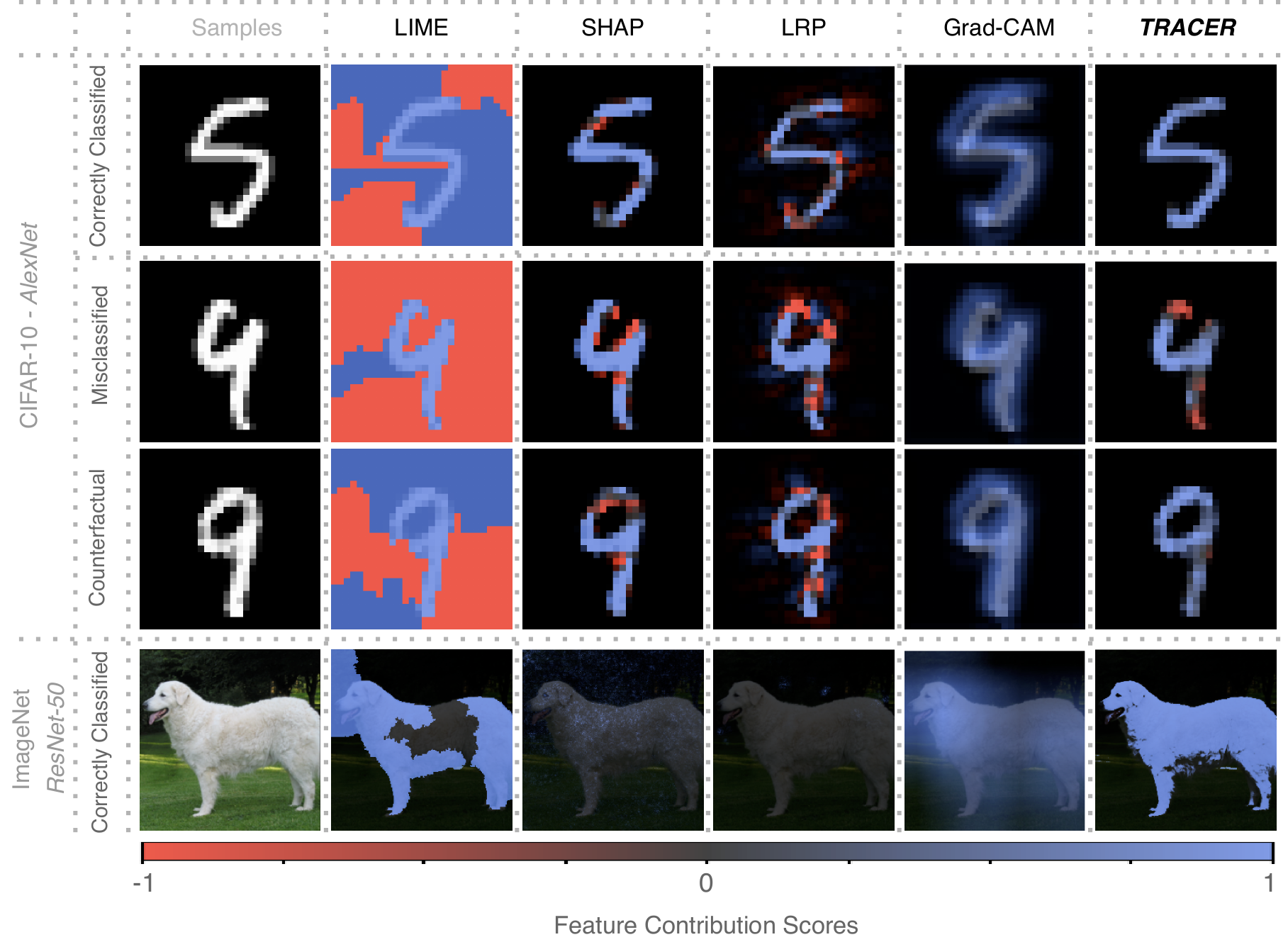}
			    \caption{{Comparison of \name results against existing explainability methods.}
			    }
			    \label{fig:comparisons}
			\end{figure}
                
                In contrast to \name,
			\begin{itemize}
				\item Every execution of \textit{LIME} produces different explanations due to its inherent stochasticity, hindering interpretability.
				\item \textit{SHAP} and \textit{LRP} explanations produce misleading results due to their sensitivity to model and dataset complexities, resulting in overly detailed or sparse attributions that do not always intuitively align with the underlying data patterns.
				\item As \textit{Grad-CAM} explanations are based on the coarse spatial resolution of the final convolutional layer of a DNN, this method often leads to highlighting broader regions rather than precise feature-level contributions to the decision-making.
                    \item {\textit{LRP} and \textit{Grad-CAM}, inherently designed for white-box DNNs, where internal model structures are accessible, face significant restrictions in terms of applicability and utility in scenarios involving black-box or proprietary models.}
			\end{itemize}

			\subsection{Tabular Datasets}\label{app:generalization_ddos}
			Transitioning from the realm of images, we further explored the efficacy of \name in the context of structured (or tabular) data. For this endeavour, we selected the CIC-IDS 2017~\citep{sharafaldin2018toward} network traffic dataset, which is representative of real-world network behaviors and patterns. This dataset poses its own set of challenges, distinct from image datasets, such as the mix of categorical and numerical attributes, the potential correlations between features, and the variance in feature scales.
			
			\begin{figure*}[h!]
			    \centering
			    \includegraphics[width=.7\linewidth]{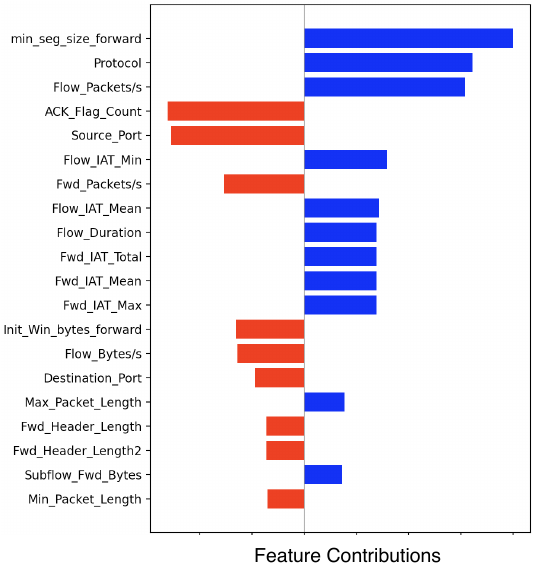}
			    \caption{Explainability of tabular datasets with \name. A sample from a Network Intrusion Detection dataset is misclassified as benign traffic rather than its correct class (DDoS attack). Negative contributions are shown in red and positive contributions in blue for the top-20 features.}
			    \label{fig:tabular}
			\end{figure*}
			
			The results presented in Figure~\ref{fig:tabular} illustrate \name's ability to provide detailed and accurate explanations beyond the image domain. For the sample explained in this figure, where a network traffic generated during a DDoS attack is considered as benign traffic by 
			{a multi-layer feed-forward neural network classifier,}		
			we observe that the features indicative of an attack negatively contribute to the decision of the classifier. Specifically, the explanations provided tell us which features were found relevant for classifying this network traffic as an attack (i.e., Source/Destination Port numbers, frequency of communication, sizes of transferred data, etc.).
			
			The clarity of the causal explanations obtained by \name for such tasks make it particularly suitable given the criticality of network intrusion detection systems in ensuring cybersecurity, where the ability to transparently understand and trust decisions can be indispensable for the practical viability of such systems.
		
		\section{Detailed results: Global Explainability}\label{app:global_explainability}
			Given the effectiveness of \name in explaining neural network decisions for individual samples, we endeavour to evaluate its potential as a global explainability tool to paint a holistic picture of the model's decision-making. To this end, rather than solely relying on global explanations, which might overlook individual nuances, we adopt an approach that aggregates local explanations to derive a global perspective. Specifically, using \name, we perform local explanations on a strategically selected subset of the dataset, aiming to capture a representative understanding of the overall characteristics. For this experiment, we selected the MNIST dataset classified using the AlexNet architecture as before. While without loss of generality, simply performing random sampling within all classes suffices for this experiment, by using clustering algorithms~\citep{settles2009active, olvera2010review} or Proximally-Connected graphs~\citep{diallo2023deciphering}, more optimal sampling policies can also be adopted to identify and select the most influential samples. Our findings for this experiment revealed several remarkable insights into the potential of \name, and into the use of AlexNet for MNIST classification.
			
			\begin{figure}[h!]
			    \centering
			    \includegraphics[width=.75\linewidth]{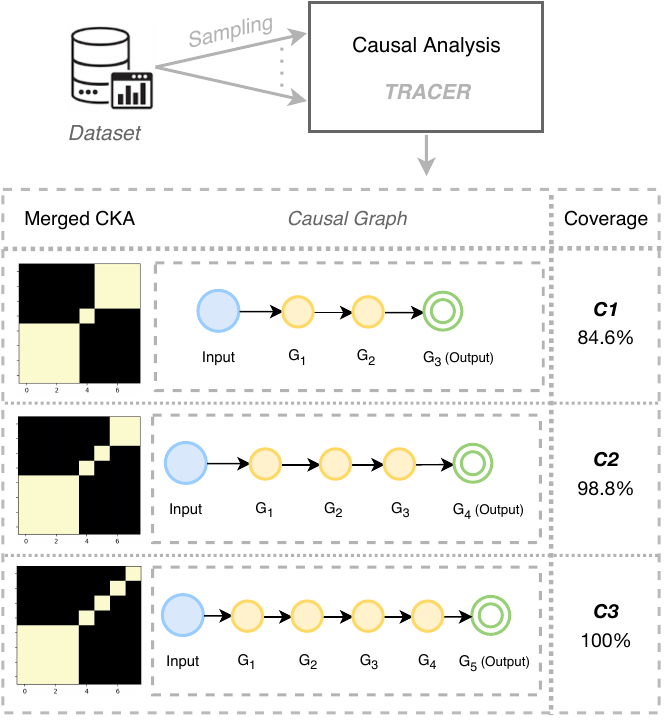}
			    \caption{Global explainability with \name -- Generalization of causal mechanisms across samples. The Coverage column indicates the percentage of analyzed samples that can be explained by distinct causal mechanisms.}
			    \label{fig:global_explanations}
			\end{figure}
			
			Specifically, as shown in Figure~\ref{fig:global_explanations}:
			\begin{enumerate}\setlength\itemsep{0em}\vspace{-1em}
				\item About 85\% of the samples could be concisely explained with a causal mechanism entailing merely 2 intermediate causal nodes. This level of generalization showcases the simplicity of the model's decision-making processes.
				\item With just one additional causal node, the causal mechanism explains ~99\% of the classifications, bringing the total to 3 intermediate causal nodes.
				\item To attain a full coverage, explaining 100\% of the classifications, the complexity increases only marginally, requiring 4 intermediate causal nodes.
			\end{enumerate}
			
			Encouraged by these insights into the causal dynamics of AlexNet's decisions on the MNIST dataset, we venture to create compressed representations of the original model. The objective is twofold: preserving the original model's accuracy while substantially reducing its computational complexity. Leveraging the knowledge distilled from \name, we craft the corresponding compressed models and train them on the identical training set as the original model (compressed models C1, C2, and C3, respectively corresponding to initial coverages of C1: 84.6\%, C2: 98.8\%, and C3: 100\%). The results, presented in Table~\ref{tab:model_compression}, show that the most compressed model achieves a staggering 99.42\% reduction in model size, while only sacrificing a negligible 0.16\% in accuracy, making it significantly more lightweight and computationally efficient.
			
			By decoding the fundamental causal interactions within neural networks, this experiment shows that \name's capacity to provide global explanations and insights can also inspire practical applications such as model compression, without compromising the integrity of the predictions. Furthermore, it is worth noting that the compressed models derived through our approach remain fully compatible with existing and well-established compression methods such as quantization and pruning, further extending their efficiency and applicability across diverse deployment scenarios.

\end{document}